\documentclass[twoside]{article}

\usepackage[accepted]{aistats2022}
\usepackage[round]{natbib}
\usepackage[utf8]{inputenc} 
\usepackage[T1]{fontenc}    
\usepackage{hyperref}       
\usepackage{url}            
\usepackage{booktabs}       
\usepackage{amsfonts}       
\usepackage{nicefrac}       
\usepackage{microtype}      
\usepackage{xcolor}         
\usepackage{graphicx}
\usepackage{multirow}
\usepackage{subcaption}
\usepackage{float}

\usepackage[ruled,vlined]{algorithm2e}
\SetKwInput{KwInput}{Input}                
\SetKwInput{KwOutput}{Output}              

\begin{document}

%

%

\twocolumn[

\aistatstitle{Dropout can Simulate Exponential Number of Models for Sample Selection Techniques}

\aistatsauthor{ Lakshya}

\aistatsaddress{ lakshya.01@samsung.com \\
Samsung R\&D Inst. Bangalore } ]

\begin{abstract}
  Following Coteaching, generally in the literature, two models are used in sample selection based approaches for training with noisy labels.  Meanwhile, it is also well known that Dropout when present in a network trains an ensemble of sub-networks. We show how to leverage this property of Dropout to train an exponential number of shared models, by training a single model with Dropout. We show how we can modify existing two model-based sample selection methodologies to use an exponential number of shared models. Not only is it more convenient to use a single model with Dropout, but this approach also combines the natural benefits of Dropout with that of training an exponential number of models, leading to improved results.
\end{abstract}

\section{Introduction}
 Noisy labels are ubiquitous in practice. For example, noise may appear due to disagreement in crowdsourcing based annotation, \cite{article1234}, or annotations carried out by computer programs on web crawled images, \cite{7953515, ratner2017data}. Consequently, it is necessary to research techniques that are robust to noisy labeling. Multiple techniques have been developed to tackle this issue. Among them, sample selection is the one that we will focus on in this paper.
 
Sample selection can be regarded as a derivative of curriculum learning, \cite{10.1145/1553374.1553380}. In sample selection techniques a curriculum is defined/learnt to select a subset of the data in each iteration of the training. MentorNet,  \cite{jiang2018mentornet}, used a single network to select a mini-batch in each iteration. Self-paced MentorNet only considered samples with a loss lower than a certain threshold for training. Coteaching,  \cite{han2018coteaching}, upgraded the MentorNet by utilizing two Networks, where the mini-batch used for training one network was decided by the loss obtained on the samples using the second network. Further, authors of Coteaching-plus , \cite{yu2019does}, argued that there should be disagreement between the two networks, which can be beneficial for the learning. Multiple techniques have been proposed since then to perform sample selection based training. However, one general trend remains among these techniques, two networks are trained for sample selection \cite{li2020dividemix, Sachdeva_2021_WACV, feng2021dmt, Wei_2020_CVPR}. 

We argue that instead of two networks, an exponential number of shared models can be utilized for sample selection. This can be achieved by utilizing Dropout, \cite{JMLR:v15:srivastava14a}, in a single network. Dropout when present in a network trains an ensemble of sub-networks, thus, it can simulate an exponential number of shared models. We can utilize this property of Dropout to transform existing two-model based approaches to utilize an exponential number of shared models by training just a single model with Dropout. This approach can combine the natural benefits of Dropout,  \cite{JMLR:v15:srivastava14a, gal2016dropout} with the benefits of training an exponential number of models for performing sample selection,  resulting in improved performance for the existing approaches when transformed to use Dropout. 

Thus, our main contributions can be listed as follows
\begin{itemize}
    \item We propose to replace the two model-based sample selection algorithms found in the literature with algorithms using an exponential number of shared models.
    \item We show how an exponential number of models can be utilized for sample selection using Dropouts in a single network.
    \item We provide empirical results by transforming existing approaches utilizing two models to use an exponential number of shared models with the help of Dropout. Our results suggest that such transformations lead to better results.
\end{itemize}

\section{Related Works}\label{related_works:section}
Various methodologies have been developed to address the problem of learning with noisy labels. Significant efforts have been invested in exploiting a noise transition matrix,  \cite{Liu_2016, hendrycks2019using, xia2020extended, li2021provably}. Efforts have also been made using graphical models, \cite{7298885, li2017learning}, and meta-learning,  \cite{ren2019learning, xu2021faster, shu2019metaweightnet, 9156647}.\cite{ciortan2021framework}  proposed using contrastive pre-training, by utilizing different pseudo-labeling and sample selection strategies, before training with a loss function. In a separate work, \cite{li2020noisy} argued that even with poor generalization, good hidden representations can be learned by the model which can be used to train a separate classifier with known correct labels.\cite{northcutt2021confident} tried to identify the label errors in the dataset by learning a joint probability distribution for noisy and clean labels under the class-conditional noise process. Meanwhile, SELF, \cite{nguyen2019self}, performs self-ensemble to filter out noisy samples from the dataset, which are then used for unsupervised loss.
 
Authors have also tried developing robust surrogate loss functions that can boost learning in the noisy setting,  \cite{patrini2017making, ma2018dimensionalitydriven, cheng2021learning, ziyin2020learning}. In particular, \cite{Wang_2019_ICCV} added a reverse cross-entropy term with the classical cross-entropy to create a symmetric cross-entropy loss. Whereas, \cite{lyu2020curriculum} proposed a curriculum loss(CL) which is a tight upper bound on the 0-1 loss. Moreover, they claimed that the CL can be used to adaptively select samples.

Another area of research is based on the early stopping criterion, for instance, \cite{liu2020earlylearning} proposed early learning regularization. Similarly, \cite{xia2021robust} argued that the parameters of a model can be divided into critical and non-critical params, which can help reduce the side effects of early learning noisy labels before early stopping.

\cite{arpit2017closer} suggested that a neural network learns easy patterns first. Multiple curriculum-based sample selection approaches have been proposed based on this observation. MentorNet by \cite{jiang2018mentornet}, Coteaching by \cite{han2018coteaching} and Coteaching-plus by \cite{yu2019does}. Similarly, JoCoR,  \cite{wei2020combating}, aims to reduce the diversity of the two models in contrast to Coteaching-plus. Recent upgrades to these models include DivideMix by \cite{li2020dividemix}, EvidentialMix by \cite{Sachdeva_2021_WACV}. They try to utilize semi-supervised learning on noisy classified labels. Another interesting approach is presented by  \cite{yi2021transform}, where a single model has been proposed for doing sample selection based on consistency of predictions by the model. However, unlike our methodology of using dropout that reduce two model networks to a single one for any two-model based approach, they rely on consistency of predictions to perform sample selection as a new approach.

\section{Using Dropout to Realize Exponential Number of Models}\label{strategy}
In this section, we describe a strategy that can be utilized to convert an existing two model-based approach to realize an exponential number of shared models.

First, we discuss the modification scheme for a model.
\begin{itemize}
    \item Generally, ML models terminate with a stack of Dense Layers. Add a Dropout unit in front of the each Dense Layer. In case there is only one or no Dense layer, add the Dropout as the pen-ultimate layer(in our experience, it worked equally well).
    \item Increase the model width for all the layers by a factor of $\frac{1}{(1-p)}$, where $p$ is the dropout probability. This step is intuitive since dropout reduces the expected width of a layer by a factor of $(1-p)$, which means that the effective width of a single network out of the exponential possibilities has reduced. 

\end{itemize}

Let's refer to a model obtained by the above modifications as DropoutNet. While, let NetA and NetB refer to two unmodified models. Note that when we forward pass with DropoutNet, a different instance of DropoutNet is obtained, based on the retained units across all the dropout layers. 

Given that, a sample selection learning algorithm(which uses NetA and NetB) can be modified as follows.
\begin{itemize}
    \item In each training iteration of the unmodified algorithm, replace NetA and NetB with different instances of DropoutNet. To get the instance for NetA, pass the mini-batch meant for NetA through the DropoutNet, similarly for NetB.
    \item Only perform backward pass through the instance corresponding to NetA.
\end{itemize}
Since, during each training iteration, different instances of the DropoutNet acts as NetA and NetB, this strategy effectively converts an existing approach to utilize an exponential number of shared models corresponding to different instances of the DropoutNet. We didn't try to perform the backward pass twice to promote implementation simplicity as when the second instance of DropoutNet is sampled, the operation graph for previous instance is removed from the memory. Moreover, since no backward pass is required for the first instance, we can direct the underlying implementation framework to not to keep track of the gradients.

\begin{figure}
\includegraphics[width=\linewidth]{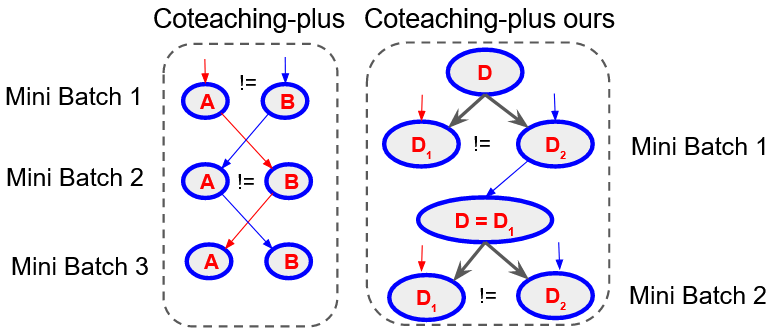}
\centering
\caption{Flow difference between Coteaching-plus-ours and Coteaching-plus. In Coteaching-plus, A chooses the small loss-data for B to train on and vice-versa. However, in Coteaching-plus-ours, two instances of D are created based on the forward pass of the mini-batch. Then, D2 chooses the small-loss data for D1 on which back propagation is performed. The updated network acts as the new D.}
\label{expteach_vs_coteach}
\end{figure}

We provide an example by modifying the Coteaching-plus algorithm using the above-mentioned steps. Let's call the resultant algorithm as Coteaching-plus-ours(in general we will use the suffix '-ours' to indicate an existing algorithm modified by our strategy). Algo-\ref{expteach_algo} shows the corresponding algorithm of Coteaching-plus-ours(symbols have the same meaning as in Coteaching-plus). Figure-\ref{expteach_vs_coteach} shows the flow difference between Coteaching-plus and Coteaching-plus-ours.

\begin{algorithm}
    \SetAlgoLined
    \KwInput{DropoutNet, training set $D$, batch size $B$, learning rate $\eta$, estimate noise rate $\tau$, epoch $E_k$ and $E_{max}$}
     \For{$e=1,..,E_{max}$}
        {
           1: Shuffle $D$  into $\frac{|D|}{B}$ mini-batches \;
           \For{$n=1,..,\frac{|D|}{B}$}
           {
                2: get the $n_{th}$ mini-batch $\bar{D}$ from $D$ \;
                \textbf{3: $w^{(1)} \gets $ DropoutNet instance by forward passing on $\bar{D}$} \;
                \textbf{4: $w^{(2)} \gets $ DropoutNet instance by re-forward passing on $\bar{D}$} \;
                5: select small-loss instance $\bar{D}'^{(2)}$ based on $w^{(2)}$ \tcp*{refer to original, \cite{yu2019does}}
                6: Update: $w^{(1)} = w^{(1)} - \eta \Delta \ell(\bar{D}'^{(2)};w^{(1)})$ \tcp*{update Dropout instance $w^{(1)}$}
           }
           7: Update: $\lambda(e)$ \tcp*{refer to original,   \cite{yu2019does}}
        }
        
        \caption{Coteaching-plus-ours}
        \label{expteach_algo}
\end{algorithm}

\section{Experimentation}\label{exp:section}
\textbf{Existing approaches.} We modified three different algorithms in our experiments, Coteaching-plus, JoCor, and DivideMix. We believe these algorithms form a good representative set of existing algorithms utilizing two model-based sample selection. 

\textbf{Datasets.} We used four different simulated noisy datasets for benchmarking, three vision-based datasets, MNIST, CIFAR-10, CIFAR-100, and one text-based dataset, NEWS. We used the same simulated noise for our experiments as done by the original approaches. Namely, Symmetric and Pair flipping or Asymmetric.

\textbf{Hyperparameters.} For all the experiments in this section, Dropout with $p=0.7$ was used. Experiments were run for 200 epochs. All the other hyper-parameters, including warm-up schedule, were kept same as the original algorithm.

\textbf{Network architecture.} For all our experiments, we used one of the following  base models(similar to Coteaching-plus).
\begin{itemize}
   \item MNIST-MLP: a 2 layer MLP with ReLU activation.
    \item CNN-small: A CNN model with 2 convolutional layers and 3 Dense layers with ReLU activation.
    \item CNN-large: A CNN model with 6 convolutional layers and 1 Dense layer with ReLU activation.
    \item NEWS-MLP: a 3 layer MLP with Softsign activation function on top of pre-trained word embeddings from GloVe, \cite{pennington-etal-2014-glove}.
\end{itemize}

Table-1 in Section-1 of supplementary materials shows the details of these networks(This table is motivated by Coteaching-plus). Table-2 in Section-1 of supplementary materials shows the details of these networks modified as per the strategy in Section-\ref{strategy}, for a dropout with $p=0.7$.


    \begin{table*}
      \caption{Average last 10 epoch test accuracy for various algorithms}
      \label{algos:table}
      \centering
      \begin{tabular}{cccccc}
        \toprule
        Algorithm & Dataset & Noise type & Noise rate & unmodified & ours\\
        \midrule
        \multirow{12}{*}{Coteaching-plus} & \multirow{3}{*}{MNIST} & sym & 0.2 & \textbf{97.75 $\pm$ 0.09} & 96.34 $\pm$ 0.14 \\
        & & sym & 0.5 & \textbf{95.88 $\pm$ 0.20} & 95.435 $\pm$ 0.19\\
        & & pairflip & 0.45 & 78.09 $\pm$ 7.78 & \textbf{87.06 $\pm$ 8.6}\\
        \cmidrule{2-6}
        & \multirow{3}{*}{CIFAR-10} & sym & 0.2 & 58.07 $\pm$ 0.88 & \textbf{58.38 $\pm$ 2.11}\\
        & & sym & 0.5 & 49.22 $\pm$ 1.04 & \textbf{53.87 $\pm$ 1.14}\\
        & & pairflip & 0.45 & 38.13 $\pm$ 1.0 & \textbf{46.72 $\pm$ 2.99} \\
        \cmidrule{2-6}
        & \multirow{3}{*}{CIFAR-100} & sym & 0.2 & 49.22 $\pm$ 0.50 & \textbf{53.47 $\pm$ 0.33}\\
        & & sym & 0.5 & 40.06 $\pm$ 0.63 & \textbf{42.72 $\pm$ 0.70}\\
        & & pairflip & 0.45 & 28.78 $\pm$ 0.32& \textbf{37.13 $\pm$ 0.88} \\
        \cmidrule{2-6}
        & \multirow{3}{*}{NEWS} & sym & 0.2 & 42.13 $\pm$ 0.51 & \textbf{46.54 $\pm$ 0.38}\\
        & & sym & 0.5 & 34.17 $\pm$ 0.65 & \textbf{39.49 $\pm$ 0.43}\\
        & & pairflip & 0.45 & 29.89 $\pm$ 0.46& \textbf{31.39 $\pm$ 0.66}\\
        
        \midrule
        \multirow{6}{*}{JoCor} & \multirow{3}{*}{CIFAR-10} & sym & 0.2 & 78.91 $\pm$ 0.19 & \textbf{83.89 $\pm$ 0.36}\\
        & & sym & 0.5 & 72.15 $\pm$ 0.58 & \textbf{76.89 $\pm$ 0.232}\\
        & & pairflip & 0.45 & 62.97 $\pm$ 0.486 & \textbf{65.16$\pm$ 1.53}\\
        \cmidrule{2-6}
        & \multirow{3}{*}{CIFAR-100} & sym & 0.2 & 44.76 $\pm$ 0.54 & \textbf{56.58 $\pm$ 0.21}\\
        & & sym & 0.5 & 36.22 $\pm$ 0.84 & \textbf{46.55 $\pm$ 0.20}\\
        & & pairflip & 0.45 & 27.18 $\pm$ 0.66& \textbf{34.00 $\pm$ 0.31}\\
        
        \midrule
        \multirow{4}{*}{DivideMix} & \multirow{3}{*}{CIFAR-10} & sym & 0.2 & 84.11 & \textbf{89.00}\\
        & & sym & 0.5 & 88.82 & \textbf{91.01}\\
        & & sym & 0.7 & 87.41 & \textbf{89.23}\\
        \cmidrule{2-6}
        & \multirow{2}{*}{CIFAR-100} & sym & 0.2 & 63.21 & \textbf{67.17}\\
        & & sym & 0.5 & 59.49 & \textbf{62.38}\\
        \bottomrule
      \end{tabular}
    \end{table*}
    

    \begin{table*}
      \caption{Average last 10 epoch test accuracy for Coteaching+Dropout and Dropout on the CIFAR-10 dataset}
      \label{no_coteach_cifar10:table}
      \centering
      \begin{tabular}{cccccc}
        \toprule
        Noise type & Noise rate & Coteaching+Dropout & Dropout\\
        \midrule
        sym & 0.2 & \textbf{58.38 $\pm$ 2.11} & 51.41 $\pm$ 1.16\\
        sym & 0.5 & \textbf{53.86 $\pm$ 1.14} & 30.37 $\pm$ 0.46\\
        pairflip & 0.45 & \textbf{46.72 $\pm$ 2.99} & 40.19 $\pm$ 1.80\\
        \bottomrule
      \end{tabular}
    \end{table*}
    

    \begin{table*}
      \caption{Average last 10 epoch test accuracy for MentorNet-ours and Coteaching-plus-ours on the CIFAR-100 dataset}
      \label{MvsC_cifar100:table}
      \centering
      \begin{tabular}{cccccc}
        \toprule
        Noise type & Noise rate & Coteaching-plus-ours & MentorNet-ours\\
        \midrule
        sym & 0.2 & \textbf{53.47 $\pm$ 0.33} & 48.96 $\pm$ 0.51\\
        sym & 0.5 & \textbf{42.72 $\pm$ 0.70} & 38.36 $\pm$ 0.62\\
        pairflip & 0.45 & \textbf{37.31 $\pm$ 0.88} & 28.17 $\pm$ 0.67\\
        \bottomrule
      \end{tabular}
    \end{table*}

\subsection{Results with Coteaching plus}
\begin{figure*}
    \begin{subfigure}{\textwidth}
      \centering
      \includegraphics[scale = 0.8]{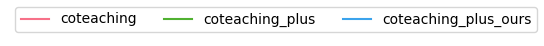}
      \label{coteach:cifar10:legend}
    \end{subfigure} 
    \\
    \begin{subfigure}{.33\textwidth}
      \centering
      \includegraphics[width=\linewidth]{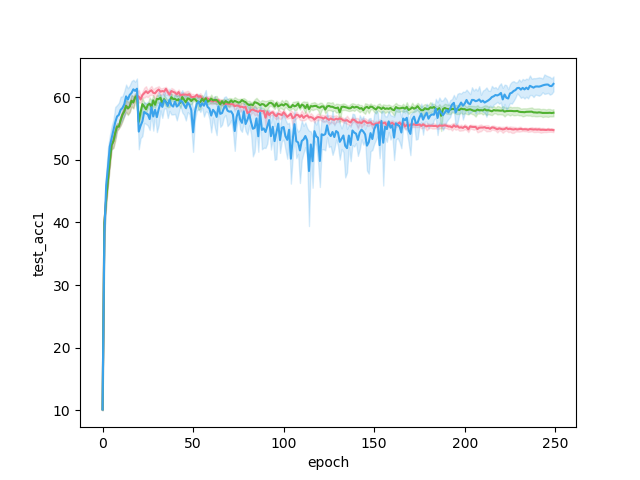}
      \caption{20\% symmetric noise}
      \label{coteach:cifar10:_0.2}
    \end{subfigure}%
    \begin{subfigure}{.33\textwidth}
      \centering
      \includegraphics[width=\linewidth]{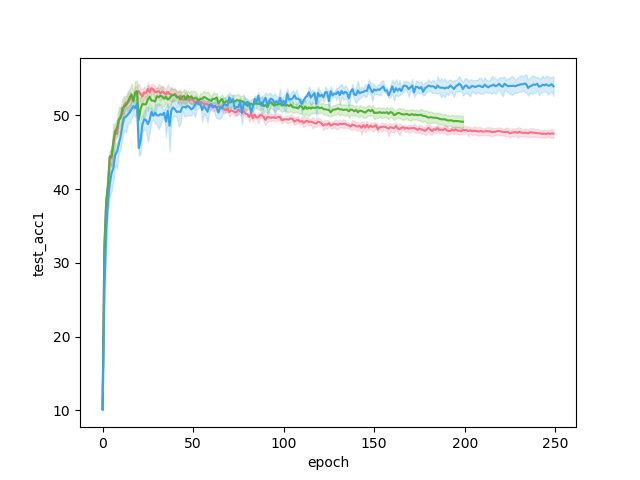}
      \caption{50\% symmetric noise}
      \label{coteach:cifar10:_0.5}
    \end{subfigure}
    \begin{subfigure}{.33\textwidth}
      \centering
      \includegraphics[width=\linewidth]{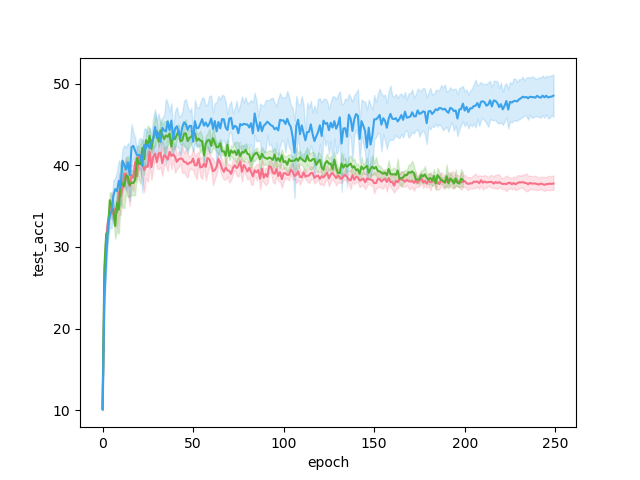}
      \caption{45\% pairflip noise}
      \label{coteach:cifar10:_0.45}
    \end{subfigure}
    \caption{Results on CIFAR-10 for experimentation with Coteaching-plus}
    \label{coteach:cifar10:fig}
    \end{figure*}

\begin{figure*}
    \begin{subfigure}{\textwidth}
      \centering
      \includegraphics[scale = 0.8]{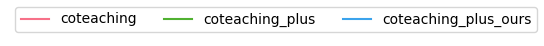}
      \label{coteach:cifar100:legend}
    \end{subfigure} 
    \\
    \begin{subfigure}{.33\textwidth}
      \centering
      \includegraphics[width=\linewidth]{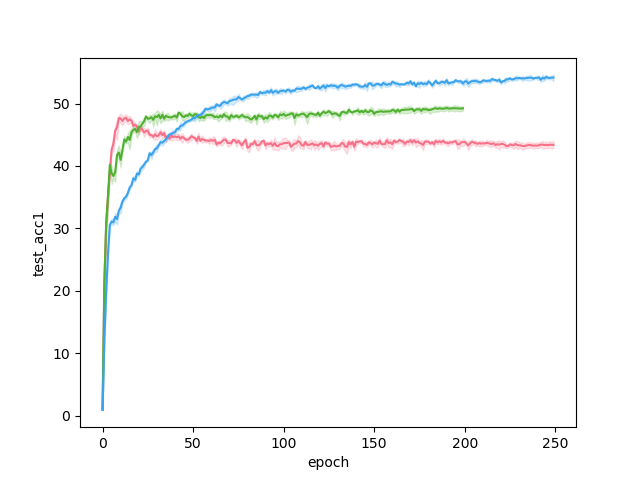}
      \caption{20\% symmetric noise}
      \label{coteach:cifar100:_0.2}
    \end{subfigure}%
    \begin{subfigure}{.33\textwidth}
      \centering
      \includegraphics[width=\linewidth]{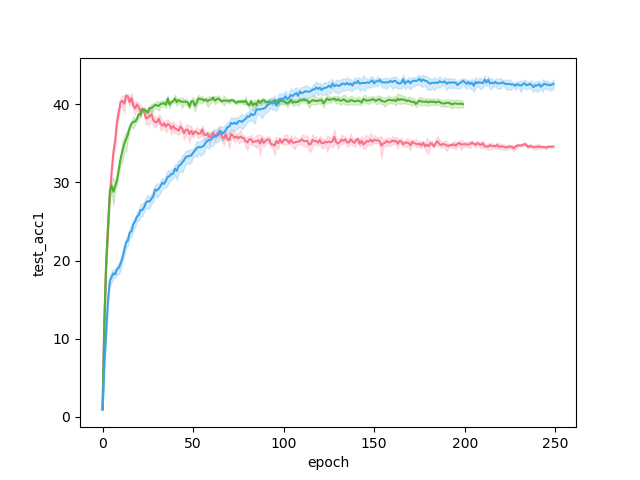}
      \caption{50\% symmetric noise}
      \label{coteach:cifar100:_0.5}
    \end{subfigure}
    \begin{subfigure}{.33\textwidth}
      \centering
      \includegraphics[width=\linewidth]{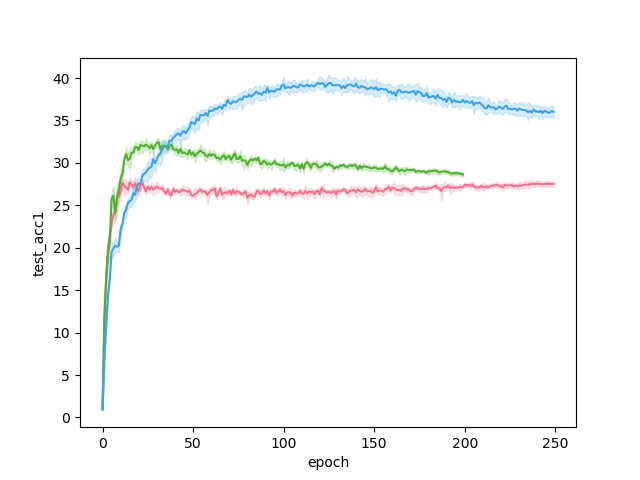}
      \caption{45\% pairflip noise}
      \label{coteach:cifar100:_0.45}
    \end{subfigure}
    \caption{Results on CIFAR-100 for experimentation with Coteaching-plus}
    \label{coteach:cifar100:fig}
    \end{figure*}

Table-\ref{algos:table} summarizes the the average last ten epoch test accuracy across five different seeds for these experiments. We used MNIST-MLP for the MNIST dataset, CNN-small for the CIFAR-10 dataset, CNN-large for the CIFAR-100 dataset, and MLP-NEWS for the NEWS dataset. Figure-\ref{coteach:cifar10:fig}, Figure-\ref{coteach:cifar100:fig} show the corresponding test accuracy vs epoch plots for CIFAR-10 and CIFAR-100 dataset(additional figures are provided in Section-2.1 of supplementary materials). The plots compare the Coteaching(unmodified), Coteaching-plus(unmodified) and Coteaching-plus-ours(modified Coteaching-plus). 

As can be seen in Table-\ref{algos:table}, except in MNIST-symmetric-0.2 and MNIST-symmetric-0.5, our approach could elevate Coteaching-plus on every other setting by values as high as 8.97\%(MNIST-pairflip-0.45), 8.59\%(CIFAR-10-pairflip-0.45), 8.35\%(CIFAR-100-pairflip-0.45).

    
\subsection{Results with JoCor}
\begin{figure*}
    \begin{subfigure}{\textwidth}
      \centering
      \includegraphics[scale = 0.8]{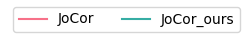}
      \label{jocor:cifar10:legend}
    \end{subfigure} 
    \\
    \begin{subfigure}{.33\textwidth}
      \centering
      \includegraphics[width=\linewidth]{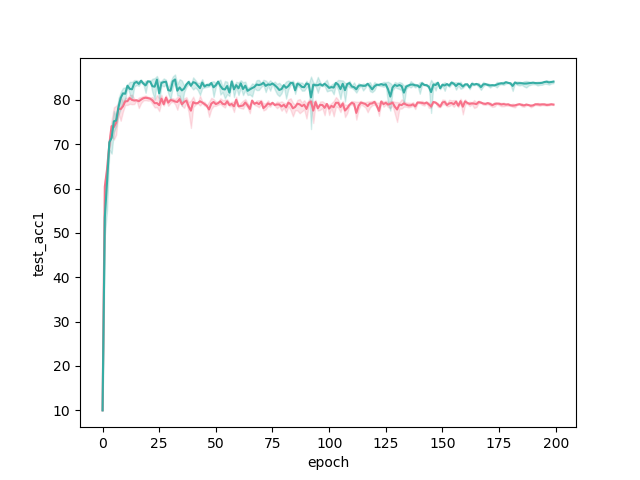}
      \caption{20\% symmetric noise}
      \label{jocor:cifar10:_0.2}
    \end{subfigure}%
    \begin{subfigure}{.33\textwidth}
      \centering
      \includegraphics[width=\linewidth]{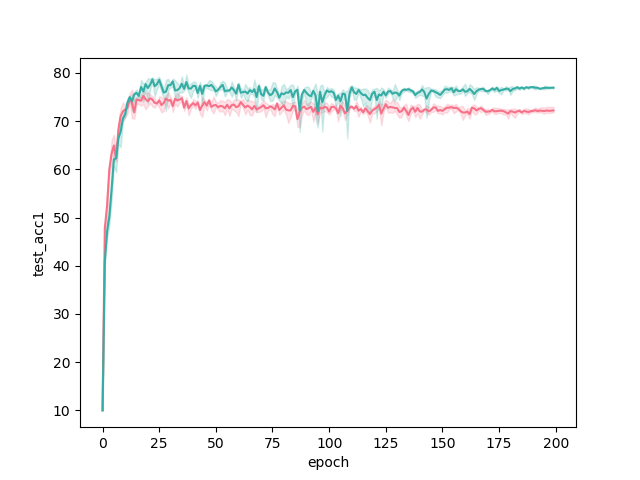}
      \caption{50\% symmetric noise}
      \label{jocor:cifar10:_0.5}
    \end{subfigure}
    \begin{subfigure}{.33\textwidth}
      \centering
      \includegraphics[width=\linewidth]{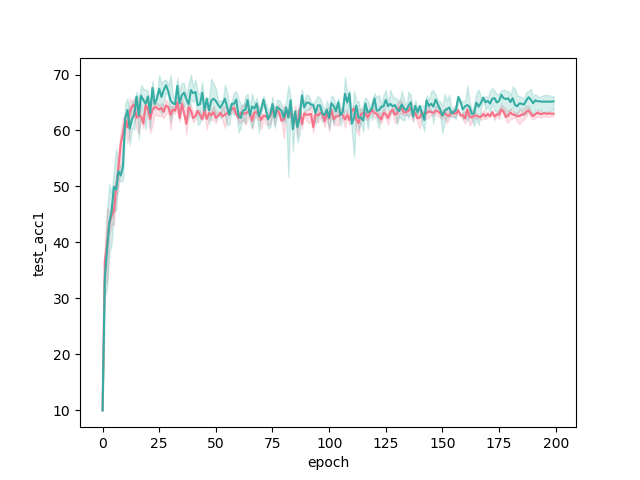}
      \caption{45\% pairflip noise}
      \label{jocor:cifar10:_0.45}
    \end{subfigure}
    \caption{Results on CIFAR-10 for experimentation with JoCor}
    \label{jocor:cifar10:fig}
    \end{figure*}

\begin{figure*}
    \begin{subfigure}{\textwidth}
      \centering
      \includegraphics[scale = 0.8]{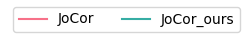}
      \label{jocor:cifar100:legend}
    \end{subfigure} 
    \\
    \begin{subfigure}{.33\textwidth}
      \centering
      \includegraphics[width=\linewidth]{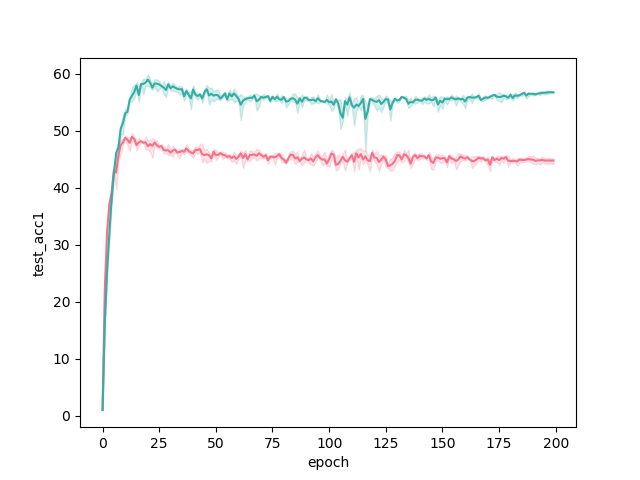}
      \caption{20\% symmetric noise}
      \label{jocor:cifar100:_0.2}
    \end{subfigure}%
    \begin{subfigure}{.33\textwidth}
      \centering
      \includegraphics[width=\linewidth]{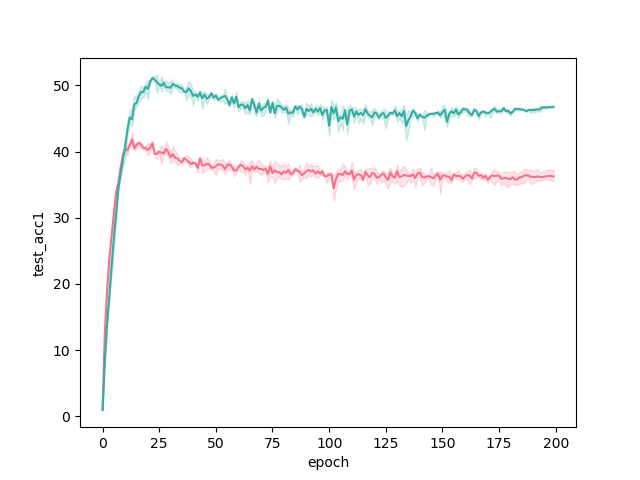}
      \caption{50\% symmetric noise}
      \label{jocor:cifar100:_0.5}
    \end{subfigure}
    \begin{subfigure}{.33\textwidth}
      \centering
      \includegraphics[width=\linewidth]{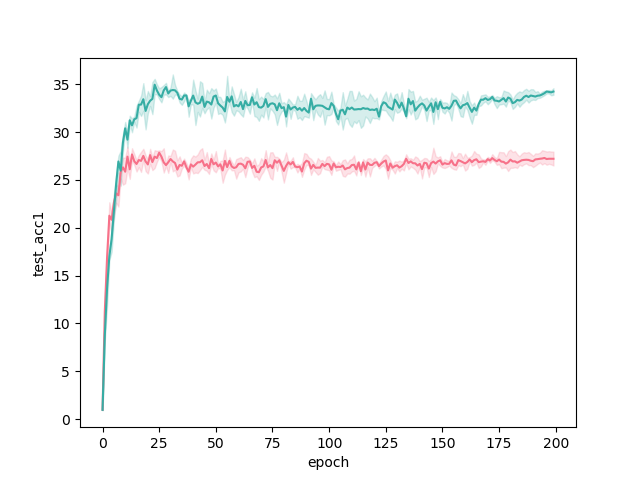}
      \caption{45\% pairflip noise}
      \label{jocor:cifar100:_0.45}
    \end{subfigure}
    \caption{Results on CIFAR-100 for experimentation with JoCor}
    \label{jocor:cifar100:fig}
    \end{figure*}

Similar to Coteaching, we calculated the average last ten epoch test accuracy across five different seeds. Table-\ref{algos:table} summarizes the results for these experiments with CIFAR-10 and CIFAR-100. For these experiments, we used CNN-large for both CIFAR-10 and CIFAR-100 datasets. Figure-\ref{jocor:cifar10:fig} and \ref{jocor:cifar100:fig} show the corresponding test accuracy vs epoch plots for CIFAR-10 and CIFAR-100 datasets.

Again, as can be seen in Table-\ref{algos:table}, our approach could elevate JoCor on every setting by values as high as 11.82\%(CIFAR-100-symmetric-0.2).

\subsection{Results with DivideMix}
\begin{figure*}
    \centering
    \begin{subfigure}{\textwidth}
      \centering
      \includegraphics[scale = 0.8]{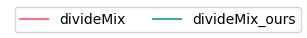}
      \label{dividemix:cifar10:legend}
    \end{subfigure} 
    \\
    \begin{subfigure}{.33\textwidth}
      \centering
      \includegraphics[width=\linewidth]{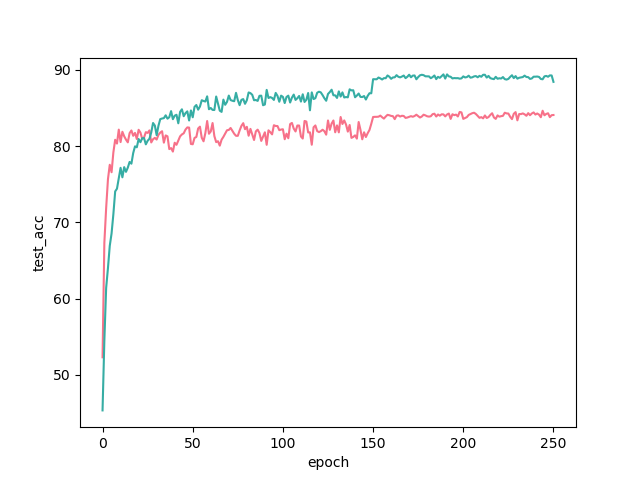}
      \caption{20\% symmetric noise}
      \label{dividemix:cifar10:_0.2}
    \end{subfigure}%
    \begin{subfigure}{.33\textwidth}
      \centering
      \includegraphics[width=\linewidth]{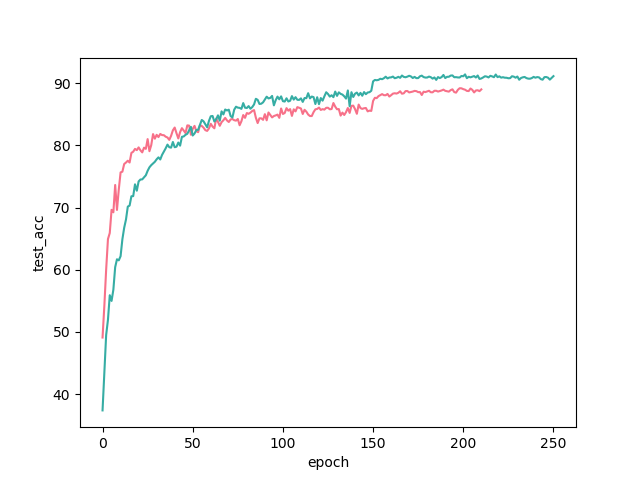}
      \caption{50\% symmetric noise}
      \label{dividemix:cifar10:_0.5}
    \end{subfigure}
    \begin{subfigure}{.33\textwidth}
      \centering
      \includegraphics[width=\linewidth]{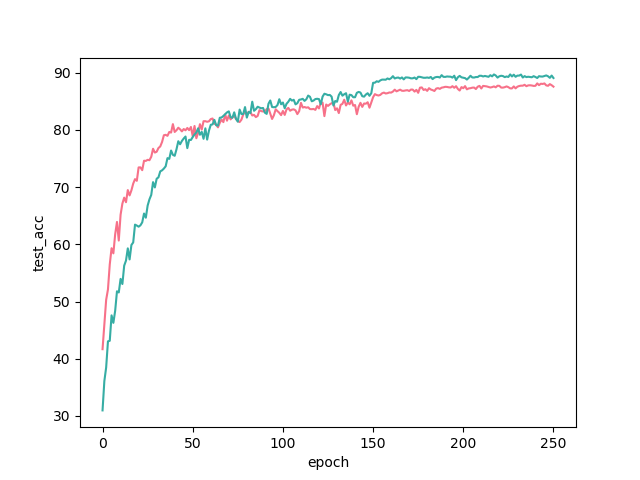}
      \caption{70\% symmetric noise}
      \label{dividemix:cifar10:_0.7}
    \end{subfigure}
    \caption{Results on CIFAR-10 for experimentation with DivideMix}
    \label{dividemix:cifar10:fig}
    \end{figure*}

\begin{figure*}
    \centering
    \begin{subfigure}{\textwidth}
      \centering
      \includegraphics[scale = 0.8]{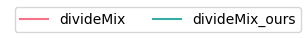}
      \label{dividemix:cifar100:legend}
    \end{subfigure} 
    \\
    \begin{subfigure}{.33\textwidth}
      \centering
      \includegraphics[width=\linewidth]{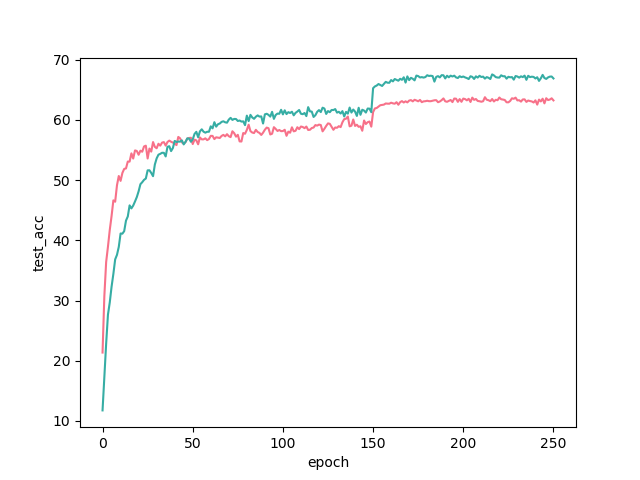}
      \caption{20\% symmetric noise}
      \label{dividemix:cifar100:_0.2}
    \end{subfigure}%
    \begin{subfigure}{.33\textwidth}
      \centering
      \includegraphics[width=\linewidth]{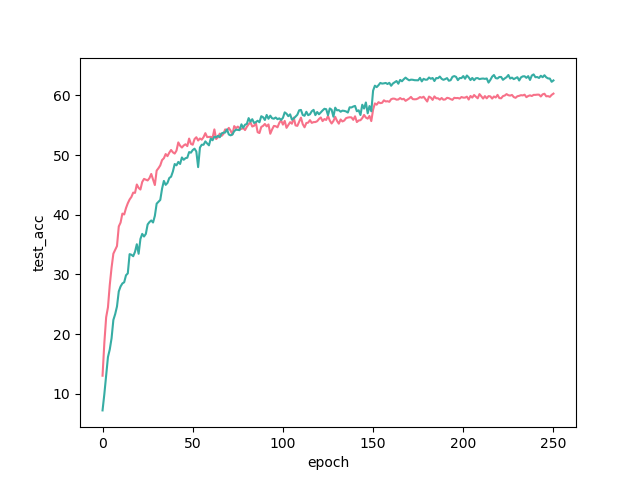}
      \caption{50\% symmetric noise}
      \label{dividemix:cifar100:_0.5}
    \end{subfigure}
    \caption{Results on CIFAR-100 for experimentation with DivideMix}
    \label{dividemix:cifar100:fig}
    \end{figure*}

For these experiments, we used only a single seed to save computation. Table-\ref{algos:table} summarizes the results for these experiments with CIFAR-10 and CIFAR-100. Similar to JoCor, we used CNN-large for both CIFAR-10 and CIFAR-100 datasets. Figure-\ref{dividemix:cifar10:fig} and \ref{dividemix:cifar100:fig} show the corresponding test accuracy vs epoch plots for CIFAR-10 and CIFAR-100 datasets.

Again, as can be seen in Table-\ref{algos:table}, our approach could elevate DivideMix on every setting by a value as high as 4.89\%(CIFAR-10-Symmetric-0.2).

\section{Additional Experiments}

\subsection{Ablation Study: Dropout without Coteaching-plus} \label{Dropout_without_coteach:section}
In these experiments, we trained a CNN-small, modified by the strategies in Section-\ref{strategy}, without any sample selection and compared it to the same model trained by Coteaching-plus-ours. The former has been referred to as the Dropout model and later as the Coteaching+Dropout model. We used the CIFAR-10 dataset and three different seeds for these experiments. Dropout with $p=0.7$ was used. Table-\ref{no_coteach_cifar10:table} summarizes the average last ten epoch test accuracy across all the seeds while Figure-3 in Section-2.2 of supplementary material shows the plot of test-accuracy vs epoch. 

By these experiments, we wanted to confirm that it is indeed the natural benefits of Dropout combined with the training algorithm that explains the test accuracy results and not just the Dropout regularization introduced to the model. From these results, it is clear that while the Dropout regularization is helpful, our approach is also able to harness the benefits of the original algorithm(Coteaching-plus).
    
\subsection{MentorNet-ours vs Coteaching-plus-ours}
In these experiments, we compared a modified CNN-large model, trained with self-paced MentorNet\footnote{self-paced MentorNet algorithm uses a single model. Hence, modification by the strategy in Section-\ref{strategy} doesn't alter the algorithm.} with the same model trained with Coteaching-plus-ours. We refer to the former as MentorNet-ours. Experiments are done on the CIFAR-100 dataset for five different seeds. Again, we used a dropout with a $p=0.7$ for these experiments. Table-\ref{MvsC_cifar100:table} summarizes the average last ten epoch test accuracy across all the seeds while Figure-4 in Section-2.3 of supplementary material shows the plot of test-accuracy vs epoch. 

By the ablation study in Section-\ref{Dropout_without_coteach:section}, we were able to confirm that our strategy is able to draw out the benefits of the Curriculum learning.  By the favorable results in this section, we can affirm that our approach can also harness the benefits of two-model based algorithms just with a single model. Since, if it couldn't the results would have been similar.

\section{Discussion}\label{discussion:section}
A notable caveat of the proposed approach is that the width of the model has to be increased such that the expected size of each instance of the modified Dropout-based model(referred to as DropoutNet in Section-\ref{strategy}) remains equivalent to the original model.

\section{Conclusion}
In this paper, we provided a strategy to elevate the existing two model-based sample selection algorithms to utilize an exponential number of shared models. We showed that this can be achieved by using a Dropout in a single model. We further provided empirical results by modifying Coteaching-plus, JoCor, and DivideMix that suggest such modification leads to better performance.



\bibliography{sample_paper}

\begin{thebibliography}{39}
\providecommand{\natexlab}[1]{#1}
\providecommand{\url}[1]{\texttt{#1}}
\expandafter\ifx\csname urlstyle\endcsname\relax
  \providecommand{\doi}[1]{doi: #1}\else
  \providecommand{\doi}{doi: \begingroup \urlstyle{rm}\Url}\fi

\bibitem[Arpit et~al.(2017)Arpit, Jastrzębski, Ballas, Krueger, Bengio,
  Kanwal, Maharaj, Fischer, Courville, Bengio, and
  Lacoste-Julien]{arpit2017closer}
Devansh Arpit, Stanisław Jastrzębski, Nicolas Ballas, David Krueger, Emmanuel
  Bengio, Maxinder~S. Kanwal, Tegan Maharaj, Asja Fischer, Aaron Courville,
  Yoshua Bengio, and Simon Lacoste-Julien.
\newblock A closer look at memorization in deep networks, 2017.

\bibitem[Bengio et~al.(2009)Bengio, Louradour, Collobert, and
  Weston]{10.1145/1553374.1553380}
Yoshua Bengio, J\'{e}r\^{o}me Louradour, Ronan Collobert, and Jason Weston.
\newblock Curriculum learning.
\newblock In \emph{Proceedings of the 26th Annual International Conference on
  Machine Learning}, ICML '09, page 41–48, New York, NY, USA, 2009.
  Association for Computing Machinery.
\newblock ISBN 9781605585161.
\newblock \doi{10.1145/1553374.1553380}.
\newblock URL \url{https://doi.org/10.1145/1553374.1553380}.

\bibitem[Cheng et~al.(2021)Cheng, Zhu, Li, Gong, Sun, and
  Liu]{cheng2021learning}
Hao Cheng, Zhaowei Zhu, Xingyu Li, Yifei Gong, Xing Sun, and Yang Liu.
\newblock Learning with instance-dependent label noise: A sample sieve
  approach, 2021.

\bibitem[Ciortan et~al.(2021)Ciortan, Dupuis, and Peel]{ciortan2021framework}
Madalina Ciortan, Romain Dupuis, and Thomas Peel.
\newblock A framework using contrastive learning for classification with noisy
  labels, 2021.

\bibitem[Feng et~al.(2021)Feng, Zhou, Gu, Tan, Cheng, Lu, Shi, and
  Ma]{feng2021dmt}
Zhengyang Feng, Qianyu Zhou, Qiqi Gu, Xin Tan, Guangliang Cheng, Xuequan Lu,
  Jianping Shi, and Lizhuang Ma.
\newblock Dmt: Dynamic mutual training for semi-supervised learning, 2021.

\bibitem[Gal and Ghahramani(2016)]{gal2016dropout}
Yarin Gal and Zoubin Ghahramani.
\newblock Dropout as a bayesian approximation: Representing model uncertainty
  in deep learning, 2016.

\bibitem[Han et~al.(2018)Han, Yao, Yu, Niu, Xu, Hu, Tsang, and
  Sugiyama]{han2018coteaching}
Bo~Han, Quanming Yao, Xingrui Yu, Gang Niu, Miao Xu, Weihua Hu, Ivor Tsang, and
  Masashi Sugiyama.
\newblock Co-teaching: Robust training of deep neural networks with extremely
  noisy labels, 2018.

\bibitem[Hendrycks et~al.(2019)Hendrycks, Mazeika, Wilson, and
  Gimpel]{hendrycks2019using}
Dan Hendrycks, Mantas Mazeika, Duncan Wilson, and Kevin Gimpel.
\newblock Using trusted data to train deep networks on labels corrupted by
  severe noise, 2019.

\bibitem[Hu et~al.(2017)Hu, Yang, Shen, Zhang, Shen, and Li]{7953515}
Mengqiu Hu, Yang Yang, Fumin Shen, Luming Zhang, Heng~Tao Shen, and Xuelong Li.
\newblock Robust web image annotation via exploring multi-facet and structural
  knowledge.
\newblock \emph{IEEE Transactions on Image Processing}, 26\penalty0
  (10):\penalty0 4871--4884, 2017.
\newblock \doi{10.1109/TIP.2017.2717185}.

\bibitem[Jiang et~al.(2018)Jiang, Zhou, Leung, Li, and
  Fei-Fei]{jiang2018mentornet}
Lu~Jiang, Zhengyuan Zhou, Thomas Leung, Li-Jia Li, and Li~Fei-Fei.
\newblock Mentornet: Learning data-driven curriculum for very deep neural
  networks on corrupted labels, 2018.

\bibitem[Li et~al.(2020{\natexlab{a}})Li, Zhang, Xu, Dickerson, and
  Ba]{li2020noisy}
Jingling Li, Mozhi Zhang, Keyulu Xu, John~P. Dickerson, and Jimmy Ba.
\newblock Noisy labels can induce good representations, 2020{\natexlab{a}}.

\bibitem[Li et~al.(2020{\natexlab{b}})Li, Socher, and Hoi]{li2020dividemix}
Junnan Li, Richard Socher, and Steven C.~H. Hoi.
\newblock Dividemix: Learning with noisy labels as semi-supervised learning,
  2020{\natexlab{b}}.

\bibitem[Li et~al.(2021)Li, Liu, Han, Niu, and Sugiyama]{li2021provably}
Xuefeng Li, Tongliang Liu, Bo~Han, Gang Niu, and Masashi Sugiyama.
\newblock Provably end-to-end label-noise learning without anchor points, 2021.

\bibitem[Li et~al.(2017)Li, Yang, Song, Cao, Luo, and Li]{li2017learning}
Yuncheng Li, Jianchao Yang, Yale Song, Liangliang Cao, Jiebo Luo, and Li-Jia
  Li.
\newblock Learning from noisy labels with distillation, 2017.

\bibitem[Liu et~al.(2020)Liu, Niles-Weed, Razavian, and
  Fernandez-Granda]{liu2020earlylearning}
Sheng Liu, Jonathan Niles-Weed, Narges Razavian, and Carlos Fernandez-Granda.
\newblock Early-learning regularization prevents memorization of noisy labels,
  2020.

\bibitem[Liu and Tao(2016)]{Liu_2016}
Tongliang Liu and Dacheng Tao.
\newblock Classification with noisy labels by importance reweighting.
\newblock \emph{IEEE Transactions on Pattern Analysis and Machine
  Intelligence}, 38\penalty0 (3):\penalty0 447–461, Mar 2016.
\newblock ISSN 2160-9292.
\newblock \doi{10.1109/tpami.2015.2456899}.
\newblock URL \url{http://dx.doi.org/10.1109/TPAMI.2015.2456899}.

\bibitem[Lyu and Tsang(2020)]{lyu2020curriculum}
Yueming Lyu and Ivor~W. Tsang.
\newblock Curriculum loss: Robust learning and generalization against label
  corruption, 2020.

\bibitem[Ma et~al.(2018)Ma, Wang, Houle, Zhou, Erfani, Xia, Wijewickrema, and
  Bailey]{ma2018dimensionalitydriven}
Xingjun Ma, Yisen Wang, Michael~E. Houle, Shuo Zhou, Sarah~M. Erfani, Shu-Tao
  Xia, Sudanthi Wijewickrema, and James Bailey.
\newblock Dimensionality-driven learning with noisy labels, 2018.

\bibitem[Nguyen et~al.(2019)Nguyen, Mummadi, Ngo, Nguyen, Beggel, and
  Brox]{nguyen2019self}
Duc~Tam Nguyen, Chaithanya~Kumar Mummadi, Thi Phuong~Nhung Ngo, Thi Hoai~Phuong
  Nguyen, Laura Beggel, and Thomas Brox.
\newblock Self: Learning to filter noisy labels with self-ensembling, 2019.

\bibitem[Northcutt et~al.(2021)Northcutt, Jiang, and
  Chuang]{northcutt2021confident}
Curtis~G. Northcutt, Lu~Jiang, and Isaac~L. Chuang.
\newblock Confident learning: Estimating uncertainty in dataset labels, 2021.

\bibitem[Patrini et~al.(2017)Patrini, Rozza, Menon, Nock, and
  Qu]{patrini2017making}
Giorgio Patrini, Alessandro Rozza, Aditya Menon, Richard Nock, and Lizhen Qu.
\newblock Making deep neural networks robust to label noise: a loss correction
  approach, 2017.

\bibitem[Pennington et~al.(2014)Pennington, Socher, and
  Manning]{pennington-etal-2014-glove}
Jeffrey Pennington, Richard Socher, and Christopher Manning.
\newblock {G}lo{V}e: Global vectors for word representation.
\newblock In \emph{Proceedings of the 2014 Conference on Empirical Methods in
  Natural Language Processing ({EMNLP})}, pages 1532--1543, Doha, Qatar,
  October 2014. Association for Computational Linguistics.
\newblock \doi{10.3115/v1/D14-1162}.
\newblock URL \url{https://www.aclweb.org/anthology/D14-1162}.

\bibitem[Ratner et~al.(2017)Ratner, Sa, Wu, Selsam, and Ré]{ratner2017data}
Alexander Ratner, Christopher~De Sa, Sen Wu, Daniel Selsam, and Christopher
  Ré.
\newblock Data programming: Creating large training sets, quickly, 2017.

\bibitem[Ren et~al.(2019)Ren, Zeng, Yang, and Urtasun]{ren2019learning}
Mengye Ren, Wenyuan Zeng, Bin Yang, and Raquel Urtasun.
\newblock Learning to reweight examples for robust deep learning, 2019.

\bibitem[Sachdeva et~al.(2021)Sachdeva, Cordeiro, Belagiannis, Reid, and
  Carneiro]{Sachdeva_2021_WACV}
Ragav Sachdeva, Filipe~R. Cordeiro, Vasileios Belagiannis, Ian Reid, and
  Gustavo Carneiro.
\newblock Evidentialmix: Learning with combined open-set and closed-set noisy
  labels.
\newblock In \emph{Proceedings of the IEEE/CVF Winter Conference on
  Applications of Computer Vision (WACV)}, pages 3607--3615, January 2021.

\bibitem[Shu et~al.(2019)Shu, Xie, Yi, Zhao, Zhou, Xu, and
  Meng]{shu2019metaweightnet}
Jun Shu, Qi~Xie, Lixuan Yi, Qian Zhao, Sanping Zhou, Zongben Xu, and Deyu Meng.
\newblock Meta-weight-net: Learning an explicit mapping for sample weighting,
  2019.

\bibitem[Srivastava et~al.(2014)Srivastava, Hinton, Krizhevsky, Sutskever, and
  Salakhutdinov]{JMLR:v15:srivastava14a}
Nitish Srivastava, Geoffrey Hinton, Alex Krizhevsky, Ilya Sutskever, and Ruslan
  Salakhutdinov.
\newblock Dropout: A simple way to prevent neural networks from overfitting.
\newblock \emph{Journal of Machine Learning Research}, 15\penalty0
  (56):\penalty0 1929--1958, 2014.
\newblock URL \url{http://jmlr.org/papers/v15/srivastava14a.html}.

\bibitem[Su et~al.(2012)Su, Deng, and Fei-Fei]{article1234}
Hao Su, J.~Deng, and L.~Fei-Fei.
\newblock Crowdsourcing annotations for visual object detection.
\newblock pages 40--46, 01 2012.

\bibitem[Wang et~al.(2019)Wang, Ma, Chen, Luo, Yi, and Bailey]{Wang_2019_ICCV}
Yisen Wang, Xingjun Ma, Zaiyi Chen, Yuan Luo, Jinfeng Yi, and James Bailey.
\newblock Symmetric cross entropy for robust learning with noisy labels.
\newblock In \emph{Proceedings of the IEEE/CVF International Conference on
  Computer Vision (ICCV)}, October 2019.

\bibitem[Wang et~al.(2020)Wang, Hu, and Hu]{9156647}
Zhen Wang, Guosheng Hu, and Qinghua Hu.
\newblock Training noise-robust deep neural networks via meta-learning.
\newblock In \emph{2020 IEEE/CVF Conference on Computer Vision and Pattern
  Recognition (CVPR)}, pages 4523--4532, 2020.
\newblock \doi{10.1109/CVPR42600.2020.00458}.

\bibitem[Wei et~al.(2020{\natexlab{a}})Wei, Feng, Chen, and An]{Wei_2020_CVPR}
Hongxin Wei, Lei Feng, Xiangyu Chen, and Bo~An.
\newblock Combating noisy labels by agreement: A joint training method with
  co-regularization.
\newblock In \emph{Proceedings of the IEEE/CVF Conference on Computer Vision
  and Pattern Recognition (CVPR)}, June 2020{\natexlab{a}}.

\bibitem[Wei et~al.(2020{\natexlab{b}})Wei, Feng, Chen, and
  An]{wei2020combating}
Hongxin Wei, Lei Feng, Xiangyu Chen, and Bo~An.
\newblock Combating noisy labels by agreement: A joint training method with
  co-regularization, 2020{\natexlab{b}}.

\bibitem[Xia et~al.(2020)Xia, Liu, Han, Wang, Deng, Li, and
  Mao]{xia2020extended}
Xiaobo Xia, Tongliang Liu, Bo~Han, Nannan Wang, Jiankang Deng, Jiatong Li, and
  Yinian Mao.
\newblock Extended t: Learning with mixed closed-set and open-set noisy labels,
  2020.

\bibitem[Xia et~al.(2021)Xia, Liu, Han, Gong, Wang, Ge, and
  Chang]{xia2021robust}
Xiaobo Xia, Tongliang Liu, Bo~Han, Chen Gong, Nannan Wang, Zongyuan Ge, and
  Yi~Chang.
\newblock Robust early-learning: Hindering the memorization of noisy labels.
\newblock In \emph{International Conference on Learning Representations}, 2021.
\newblock URL \url{https://openreview.net/forum?id=Eql5b1_hTE4}.

\bibitem[Xiao et~al.(2015)Xiao, Xia, Yang, Huang, and Wang]{7298885}
Tong Xiao, Tian Xia, Yi~Yang, Chang Huang, and Xiaogang Wang.
\newblock Learning from massive noisy labeled data for image classification.
\newblock In \emph{2015 IEEE Conference on Computer Vision and Pattern
  Recognition (CVPR)}, pages 2691--2699, 2015.
\newblock \doi{10.1109/CVPR.2015.7298885}.

\bibitem[Xu et~al.(2021)Xu, Zhu, Jiang, and Yang]{xu2021faster}
Youjiang Xu, Linchao Zhu, Lu~Jiang, and Yi~Yang.
\newblock Faster meta update strategy for noise-robust deep learning, 2021.

\bibitem[Yi and Huang(2021)]{yi2021transform}
Rumeng Yi and Yaping Huang.
\newblock Transform consistency for learning with noisy labels, 2021.

\bibitem[Yu et~al.(2019)Yu, Han, Yao, Niu, Tsang, and Sugiyama]{yu2019does}
Xingrui Yu, Bo~Han, Jiangchao Yao, Gang Niu, Ivor~W. Tsang, and Masashi
  Sugiyama.
\newblock How does disagreement help generalization against label corruption?,
  2019.

\bibitem[Ziyin et~al.(2020)Ziyin, Chen, Wang, Liang, Salakhutdinov, Morency,
  and Ueda]{ziyin2020learning}
Liu Ziyin, Blair Chen, Ru~Wang, Paul~Pu Liang, Ruslan Salakhutdinov,
  Louis-Philippe Morency, and Masahito Ueda.
\newblock Learning not to learn in the presence of noisy labels, 2020.

\end{thebibliography}
\end{document}